\documentclass[10pt,twocolumn,letterpaper]{article}

\usepackage[pagenumbers]{cvpr}      
\usepackage{pifont}
\usepackage{multirow}
\usepackage{comment}
\usepackage{rotating}
\usepackage{makecell}
\usepackage{tabularx}
\usepackage{float}

\usepackage[dvipsnames]{xcolor}

\definecolor{cvprblue}{rgb}{0.21,0.49,0.74}
\usepackage[pagebackref,breaklinks,colorlinks,citecolor=cvprblue]{hyperref}

\def\paperID{2644} 
\def\confName{CVPR}
\def\confYear{2024}

\title{SEED-Bench-2-Plus: Benchmarking Multimodal Large Language Models\\ with Text-Rich Visual Comprehension}

\author{
Bohao Li$^{3,1}$ \and
Yuying Ge$^{1\dagger}$ \and
Yi Chen$^{1}$ \and
Yixiao Ge$^{1,2}$ \and
Ruimao Zhang$^{3\dagger}$ \and
Ying Shan$^{1,2}$ \and
\\
$^{1}$Tencent AI Lab \\
$^{2}$ARC Lab, Tencent PCG \\
$^{3}$School of Data Science, The Chinese University of HongKong, Shenzhen \\
}

\begin{document}

\twocolumn[{%
\renewcommand\twocolumn[1][]{#1}%
    \maketitle
}]
\renewcommand{\thefootnote}{\fnsymbol{footnote}}
   \footnotetext[1]{This technical report is an extension to SEED-Bench-2~\cite{li2023seed2}.}
   \footnotetext[2]{Correspondence Author.}

\begin{table*}[]
    \centering
    \caption{Comparisons between existing MLLM benchmarks. ``H/G Evaluation'' denotes whether human or GPT is used for evaluation. ``\#Models'' denotes the number of evaluated model.}\label{tab:benchmark_compare}
    {\small
    \resizebox{\textwidth}{!}{
    \begin{tabular}{cccccccc}
         \toprule
             {\multirow{2}{*}{Benchmark}} & \multirow{1}{*}{\makecell{Customized\\Question}} & \multirow{1}{*}{\makecell{Text-Rich\\Data}} & \multirow{1}{*}{\makecell{Text-Rich\\Scenes}} & \multirow{1}{*}{\makecell{\#Answer\\Annotation}} & \multirow{1}{*}{ \makecell{Answer\\Type}} & \multirow{1}{*}{\makecell{H/G\\Evaluation}} & {\multirow{2}{*}{\#Models}}\\
         \\
         \midrule
         LLaVA-Bench~\cite{liu2023visual_llava} & \ding{51} & \ding{55} & - & - & free-form & GPT & 4\\
         MME~\cite{fu2023mme} & \ding{51}  & \ding{55} & - & - &Y/N & N/A & 10\\
         M3Exam~\cite{zhang2023m3exam} &\ding{51}  & \ding{55} & - & - & A/B/C/D & N/A & 7\\
         LAMM~\cite{yin2023lamm} & \ding{55} & \ding{55} & - & - & free-form & GPT & 4\\
         LVLM-eHub~\cite{xu2023lvlm} & \ding{55} & \ding{55} & - & - & free-form & Human &8\\
         MMBench~\cite{liu2023mmbench} & \ding{55} & \ding{55} & - &- & free-form & GPT &14\\
         VisIT-Bench~\cite{bitton2023visitbench} & \ding{51} & \ding{55} & - & - &free-form &Human/GPT &14\\
         MM-VET~\cite{yu2023mmvet}  &\ding{51} & \ding{55} & - & - &free-form & GPT &9 \\
         Touchstone~\cite{bai2023touchstone} &\ding{51} & \ding{55} & - & - & free-form & GPT & 7\\
         Q-bench~\cite{wu2023qbench} &\ding{51} & \ding{55} & - &- &Y/N \& free-form & N/A & 15\\
         SEED-Bench-1~\cite{li2023seed} &\ding{51} & \ding{55} & - & - & A/B/C/D & N/A & 18\\
         SEED-Bench-2~\cite{li2023seed2} & \ding{51} & \ding{55} & - & - & A/B/C/D & N/A & 23\\
         OCR-Bench~\cite{liu2023ocrbench} & \ding{55} & \ding{51} & 28 & - &free-form & N/A & 6 \\
         CONTEXTUAL~\cite{wadhawan2024contextual}  &\ding{51} & \ding{51} & 8 & 506 & free-form & GPT & 13\\
         MathVista~\cite{lu2023mathvista} &\ding{51} & \ding{51} & 19 & 735 & A/B/C/D \& free-form &N/A & 12\\
         MMMU~\cite{yue2023mmmu} &\ding{51} & \ding{51} & 30 &11.5K & A/B/C/D \& free-form & N/A & 23\\
         SEED-Bench-2-Plus(Ours) & \ding{51} & \ding{51} & 63 & 2.3K & A/B/C/D & N/A & 34\\
         \bottomrule
    \vspace{4pt}
    \end{tabular}
    }
   }
\end{table*}

\begin{abstract}
Comprehending text-rich visual content is paramount for the practical application of Multimodal Large Language Models (MLLMs), since text-rich scenarios are ubiquitous in the real world, which are 
characterized by the presence of extensive texts embedded within images. Recently, the advent of MLLMs with impressive versatility has raised the bar for what we can expect from MLLMs. However, their proficiency in text-rich scenarios has yet to be comprehensively and objectively assessed, since current MLLM benchmarks primarily focus on evaluating general visual comprehension. In this work, we introduce SEED-Bench-2-Plus, a benchmark specifically designed for evaluating \textbf{text-rich visual comprehension} of MLLMs. Our benchmark comprises 2.3K multiple-choice questions with precise human annotations, spanning three broad categories: Charts, Maps, and Webs, each of which covers a wide spectrum of text-rich scenarios in the real world. These categories, due to their inherent complexity and diversity, effectively simulate real-world text-rich environments. We further conduct a thorough evaluation involving 34 prominent MLLMs (including GPT-4V, Gemini-Pro-Vision and Claude-3-Opus) and emphasize the current limitations of MLLMs in text-rich visual comprehension.
We hope that our work can serve as a valuable addition to existing MLLM benchmarks, providing insightful observations and inspiring further research in the area of text-rich visual comprehension with MLLMs. The dataset and evaluation code can be accessed at \href{https://github.com/AILab-CVC/SEED-Bench}{\color{magenta}https://github.com/AILab-CVC/SEED-Bench.}

\end{abstract}
    
\section{Introduction}

\begin{figure}
    \centering
    \begin{subfigure}{0.5\textwidth}
        \centering
        \includegraphics[width=\linewidth]{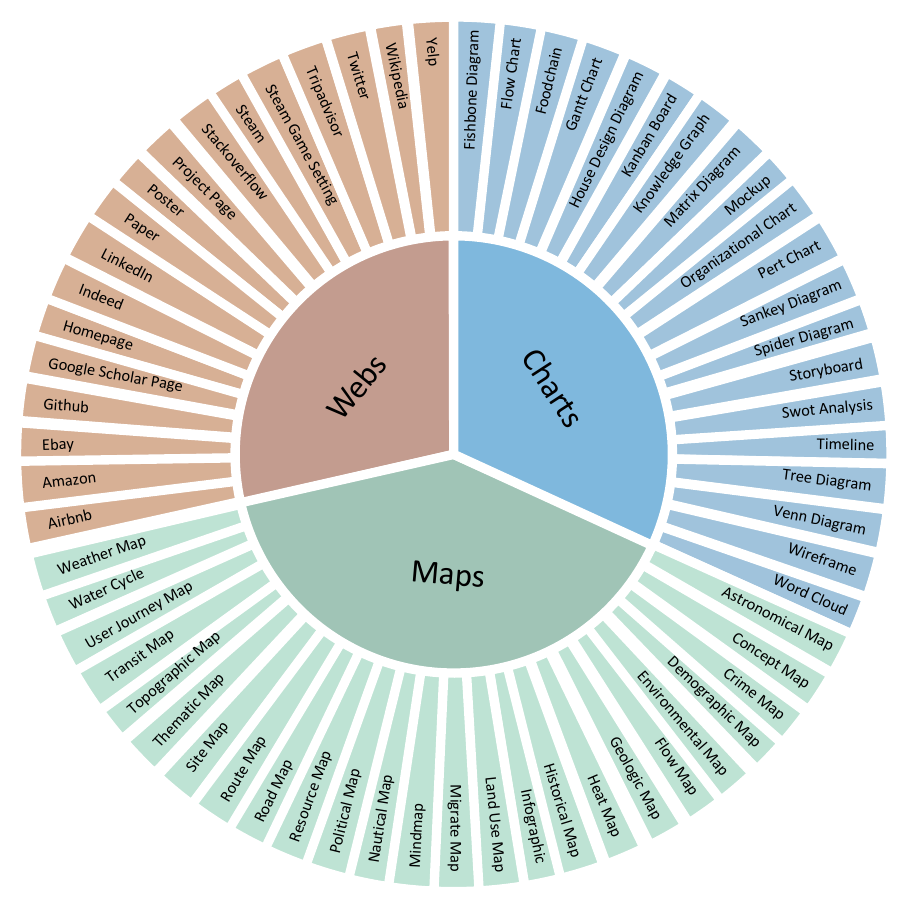}
    \end{subfigure}
    \vspace{2pt}
    \caption{Overview of 63 data types within three major categories including Charts, Maps, and Webs in SEED-Bench-2-Plus, which encompasses an extensive range of \textbf{text-rich scenarios} for evaluating visual comprehension of MLLMs.}
    \label{fig:text_rich_overview}
\end{figure}

In recent years, through leveraging the strong generality of Large Language Models (LLMs)~\cite{chung2022scaling_flant5, openai2023gpt4, ChatGPT, vicuna, touvron2023llama}, Multimodal Large Language Models (MLLMs)~\cite{li2023blip2, zhu2023minigpt4, team2023gemini, liu2023visual_llava, ye2023mplugowl, dai2023instructblip, li2023otter, gong2023multimodalgpt, su2023pandagpt, peng2023kosmos, li2023videochat, maaz2023videochatgpt, luo2023valley, peng2023kosmos, bai2023qwen, liu2023llava1.5, laurencon2023obelics, zhang2023internlm, ge2023making, 2023GPT4VisionSC} have showcased remarkable capabilities in comprehending multimodal data, which aim to mimic human-like understanding through multimodal perception and reasoning. To realize the practical applications of MLLMs in the real world, a significant challenge lies in comprehending text-rich visual data, which is pervasive in various contexts with images intertwined with texts. An MLLM capable of comprehensively comprehending text-rich scenarios should be able to interpret texts, understand visual content, and discern the interactions between textual and visual contexts.

\begin{figure*}
    \centering
    \includegraphics[width=0.95\textwidth]{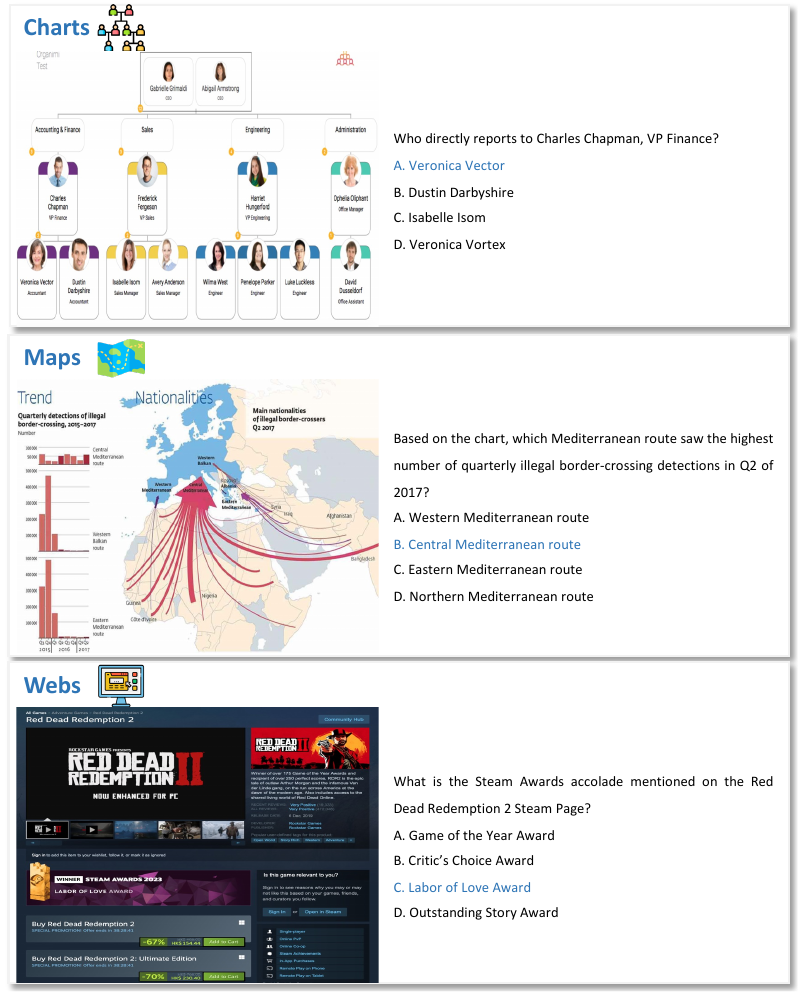}
    \caption{Data samples in SEED-Bench-2-Plus spanning three broad categories: Charts,
Maps, and Webs. These categories cover a wide spectrum of multiple-choice questions in text-rich scenarios, with ground truth options derived from human annotations.}
    \label{fig:text_rich_example}
        \vspace{2pt}
\end{figure*}

With the emergence of increasingly powerful and versatile MLLMs, such as GPT-4V~\cite{2023GPT4VisionSC}, Gemini-Pro-Vision~\cite{team2023gemini}, and Claude-3-Opus~\cite{claude3family}, it naturally raises the question: \textit{How do these models perform in text-rich scenarios?} Although recent benchmarks~\citep{liu2023mmbench, fu2023mme, li2023seed, li2023seed2} are specifically designed to evaluate MLLMs, their primary focus is on general visual comprehension (\textit{e.g.}, images in different domains), leaving a significant gap in a comprehensive and objective evaluation of MLLM in text-rich contexts. 

In this work, we introduce SEED-Bench-2-Plus, a comprehensive benchmark designed specifically to assess MLLMs' performance in comprehending text-rich visual data, which covers a wide spectrum range of text-rich scenarios in the real world. Specially, we meticulously craft 2.3K multiple-choice questions spanning three broad categories including Charts, Maps and Charts, as shown in Figure~\ref{fig:text_rich_example}. The broad categories are further divided into 63 specific types as shown in Figure~\ref{fig:text_rich_overview}, to capture a more granular view of the challenges presented by text-rich visual comprehension. Each question in our benchmark is answered by human annotators, ensuring the accuracy and reliability of the ground-truth answer.

We further conduct extensive evaluation, encompassing 34 prominent MLLMs, including GPT-4V, Gemini-Pro-Vision, and Claude-3-Opus, which reveals critical insights into the current limitations and strengths of these models in comprehending text-rich information. Our findings underline several key observations: the inherent complexity of text-rich visual data, the varying difficulty levels across different data types, and the performance disparities among leading MLLMs. 

Through this evaluation, we aim not only to benchmark current MLLM performance but also to catalyze further research in enhancing MLLMs' proficiency of multimodal comprehension in text-rich scenarios. Serving as a valuable supplement to Seed-Bench-2~\cite{li2023seed2}, both the dataset and the evaluation code of Seed-Bench-2-Plus are made publicly available. We also consistently maintain a leaderboard to facilitate ongoing contributions to this area.

\begin{figure}
    \includegraphics[width=0.48\textwidth]{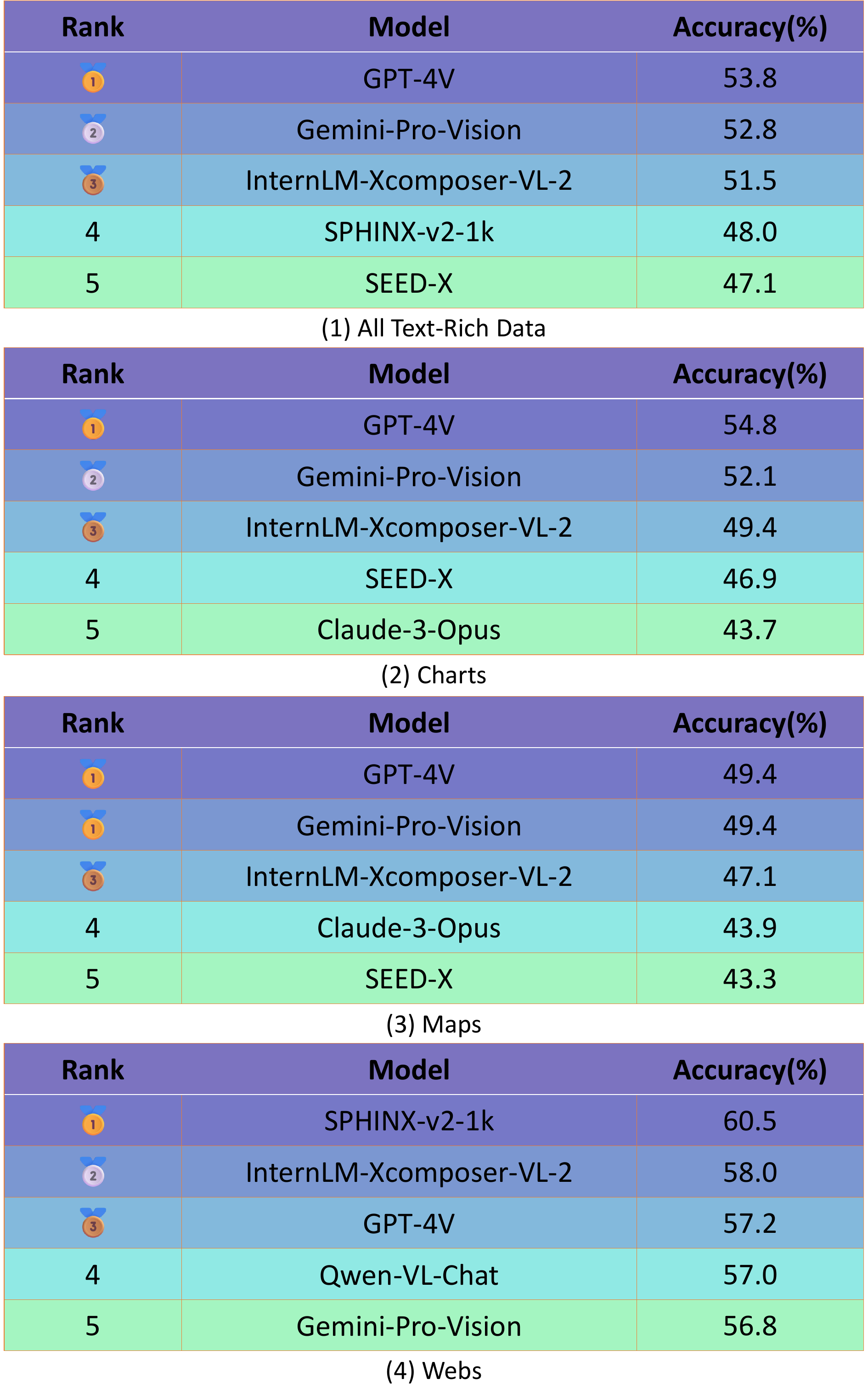}
    \caption{Leaderboard of SEED-Bench-2-Plus.}
    \label{fig:leaderboard}
        \vspace{-12pt}
\end{figure}

\section{Related Work}\label{sec:related_work}
\noindent\textbf{Multimodal Large Language Models.}
Following the remarkable success of Large Language Models (LLM)~\cite{chung2022scaling_flant5,touvron2023llama,vicuna}, recent studies work on generative Multimodal Large Language Models (MLLMs)~\cite{li2023blip2, zhu2023minigpt4, liu2023visual_llava, ye2023mplugowl, dai2023instructblip, li2023otter, gong2023multimodalgpt, su2023pandagpt,peng2023kosmos, bai2023qwen, liu2023llava1.5, laurencon2023obelics, zhang2023internlm} aim to enhance multimodal comprehension by aligning the visual features of pre-trained image encoders with LLMs on image-text datasets. Some research~\cite{li2023videochat,maaz2023videochatgpt,luo2023valley} has further incorporated video inputs, leveraging the vast capabilities of LLMs for video understanding tasks. Recent work~\cite{sun2023emu, yu2023scaling, ge2023planting, ge2023making, wu2023nextgpt, dong2023dreamllm} take significant strides in equipping MLLMs with the capacity for generating images beyond texts. However, their capability in text-rich scenarios has yet to be comprehensively and objectively assessed, which is the primary motivation of SEED-Bench-2-Plus.

\noindent\textbf{Benchmarks for Multimodal Large Language Models.}
In tandem with the rapid development of Multimodal Large Language Models (MLLMs), several concurrent studies~\cite{fu2023mme,yin2023lamm,xu2023lvlm,liu2023mmbench,bai2023touchstone} have proposed various benchmarks for evaluating MLLMs. 
For instance, GVT~\cite{wang2023gvt} constructs a benchmark by aggregating two semantic-level understanding tasks (VQA and Image Captioning) and two fine-grained tasks (Object Counting and Multi-class Identification). However, its evaluation is limited to specific aspects of visual understanding. LVLM-eHub~\cite{xu2023lvlm} combines multiple existing computer vision benchmarks and develops an online platform where two models are prompted to answer a question related to an image, and human annotators are employed to compare the models' predictions. The involvement of human annotators during evaluation not only introduces bias but also incurs significant costs. LLaVA-Bench~\cite{liu2023visual_llava}, LAMM~\cite{yin2023lamm}, and Touchstone~\cite{bai2023touchstone} utilize GPT to evaluate the relevance and accuracy of answers in relation to the ground truth. The reliance on entity extraction and GPT metric can affect the accuracy and reliability of the evaluation. MME~\cite{fu2023mme} and MMBench~\cite{liu2023mmbench} aim to enhance the objective evaluation of MLLMs by constructing 2194 True/False Questions and 2974 Multiple Choice Questions across various ability dimensions, respectively. MMMU~\cite{yue2023mmmu} generates numerous college-level multi-discipline QAs to evaluate MLLMs' knowledgeability.
In SEED-Bench-2-Plus, we focus on evaluating MLLMs’ performance in comprehending text-rich visual
data, which covers a wide spectrum range of text-rich scenarios in the real world.

\section{SEED-Bench-2-Plus}
\begin{table*}[]
    \centering
    \caption{Evaluation results of various MLLMs in SEED-Bench-2-Plus. The best (second best) is in bold (underline).}\label{tab:performance}
    \vspace{3pt}
    {\small
    \resizebox{0.8\textwidth}{!}{
    \begin{tabular}{cccccc}
         \toprule
         {\multirow{2}{*}{Model}} & \multirow{2}{*}{Language Model}  &  \multicolumn{4}{c}{Text-Rich Data}   \\
         \cmidrule(lr){3-6}
         & &\multirow{1}{*}{Average} & \multirow{1}{*}{Charts} & \multirow{1}{*}{Maps} & \multirow{1}{*}{Webs} \\
         \midrule
         BLIP-2~\cite{li2023blip2} &Flan-T5-XL & 29.8 & 33.6 & 30.2 & 25.6\\
         InstructBLIP~\cite{dai2023instructblip} &Flan-T5-XL & 29.2 & 31.7 & 28.7 & 27.2 \\
         InstructBLIP Vicuna~\cite{dai2023instructblip} &Vicuna-7B & 30.9 & 30.0 & 32.7 & 29.9 \\
         LLaVA~\cite{liu2023visual_llava} &LLaMA-7B & 30.1 & 29.9 & 29.0 & 31.3 \\
         MiniGPT-4~\cite{zhu2023minigpt4} &Vicuna-7B & 30.2 & 30.5 & 30.4 & 29.7\\
         VPGTrans~\cite{2023vpgtrans} &LLaMA-7B &30.3 & 30.0 & 31.3 & 29.6 \\
         MultiModal-GPT~\cite{gong2023multimodalgpt} &Vicuna-7B &31.7 & 30.5 & 32.7 & 32.0 \\
         Otter~\cite{li2023otter} &LLaMA-7B &31.3 & 29.5 & 32.3 & 32.0 \\
         OpenFlamingo~\cite{openflamingo} &LLaMA-7B &31.7 & 30.5 & 32.7 & 32.0 \\
         LLaMA-Adapter V2~\cite{gao2023llamaadapterv2} &LLaMA-7B &30.6 & 29.9 & 30.8 & 31.1 \\
         GVT~\cite{wang2023gvt} &Vicuna-7B &29.7 & 29.3 & 30.2 & 29.7\\
         mPLUG-Owl~\cite{ye2023mplugowl} &LLaMA-7B &31.8 & 30.6 & 33.5 & 31.1\\
         Qwen-VL~\cite{bai2023qwen} & Qwen-7B &37.0 & 38.2 & 37.0 & 55.9 \\
         Qwen-VL-Chat~\cite{bai2023qwen} & Qwen-7B &43.4 & 37.3 & 35.9 & 57.0 \\
         LLaVA-1.5~\cite{liu2023llava1.5} &Vicuna-7B &36.8 & 36.5 & 35.1 & 38.8 \\
         IDEFICS-9B-Instruct~\cite{laurencon2023obelics} & LLaMA-7B &32.1 & 31.0 & 31.8 & 33.5\\
         InternLM-Xcomposer-VL~\cite{zhang2023internlm} & InternLM-7B &40.6 & 39.9 & 39.0 & 43.0 \\
         VideoChat~\cite{li2023videochat} &Vicuna-7B &28.6 & 27.8 & 29.7 & 28.3 \\
         Video-ChatGPT~\cite{maaz2023videochatgpt} &LLaMA-7B &29.8 & 29.9 & 29.0 & 30.5 \\
         Valley~\cite{luo2023valley} &LLaMA-13B &29.2 & 29.1 & 27.4 & 31.1 \\
         Emu~\cite{sun2023emu} & LLaMA-13B & 33.5 & 32.4 & 34.2 & 34.0\\
         NExt-GPT~\cite{wu2023nextgpt} & Vicuna-7B & 26.2 & 26.3 & 26.6 & 25.7\\
         SEED-LLaMA~\cite{ge2023making} & LLaMA2-Chat-13B & 33.7 & 32.5 & 35.7 & 33.1\\
         CogVLM~\cite{wang2023cogvlm} & Vicuna-7B &33.4 & 32.6 & 34.1 & 33.5\\
         InternLM-Xcomposer-VL2~\cite{dong2024internlm2} & InternLM2-7B & 51.5 & 49.4 & 47.1 & \underline{58.0} \\
         InternLM-Xcomposer-VL2-4bit~\cite{dong2024internlm2} & InternLM2-7B &37.6 & 37.4 & 38.8 & 36.7 \\
         LLaVA-Next~\cite{liu2024llava_next} & Vicuna-7B & 36.8 & 36.4 & 34.0 & 39.9 \\
         Yi-VL~\cite{yi_vl_6b} &Yi-6B  &34.8 & 32.4 & 34.6 & 37.5\\
         SPHINX-v2-1k~\cite{gao2024sphinx} & LLaMA2-13B & 48.0 & 41.7 & 41.9 & \bf{60.5} \\
         mPLUG-Owl2~\cite{ye2023mplug2} & LLaMA2-7B &33.4 & 33.5 & 32.6 & 34.0 \\
         SEED-X~\cite{ge2024seedx} & LLaMA2-13B & 47.1 & 46.9 & 43.3 & 52.6 \\
         GPT-4V~\cite{2023GPT4VisionSC} & - &\bf{53.8} & \bf{54.8} & \bf{49.4} & {57.2} \\
         Gemini-Pro-Vision~\cite{team2023gemini} & - & \underline{52.8} & \underline{52.1} & \bf{49.4} & {56.8} \\
         Claude-3-Opus~\cite{claude3family} & - & 44.2 & 43.7 & 43.9 & 45.1\\
         \bottomrule
         \vspace{2pt}
    \end{tabular}
    }
   }
\end{table*}

\vspace{10pt}
\begin{figure*}
    \includegraphics[width=0.95\textwidth]{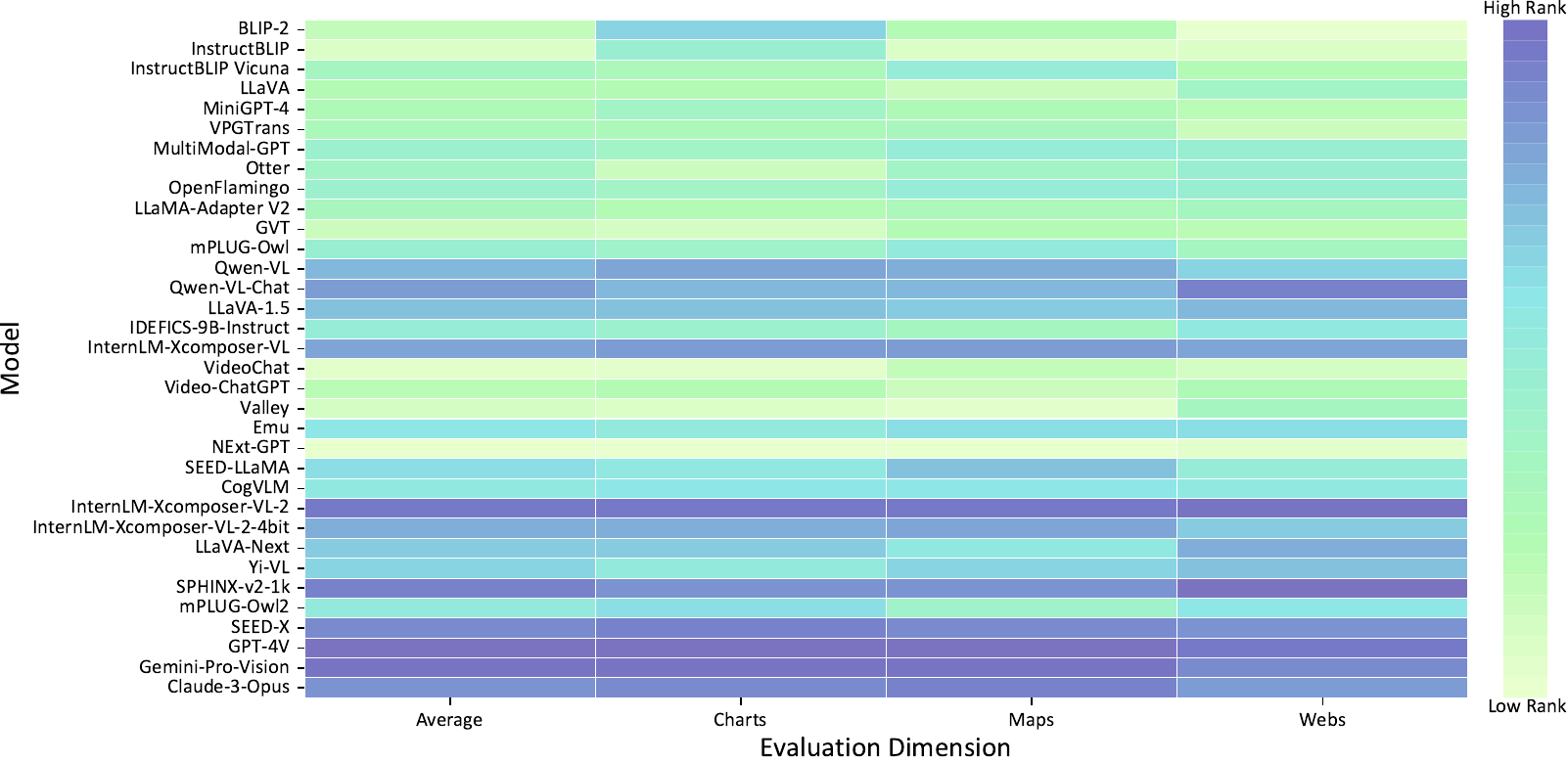}
    \caption{Illustration of each model's performance across different categories in SEED-Bench-2-Plus with text-rich scenarios, where darker colors represent higher ranks.}
    \label{fig:rank}
    \vspace{10pt}
\end{figure*}

\begin{table*}[]
    \centering
    \caption{Evaluation results of various MLLMs in different types of ``Charts'' (Part 1) in SEED-Bench-2-plus. The best (second best) is in bold (underline).} \label{tab:chart_performance_part1}
    \vspace{3pt}
    {
    \resizebox{\textwidth}{!}{
    \begin{tabular}{cccccccccccc}
         \toprule
         \multirow{1}{*}{Model} & \multirow{1}{*}{Language Model}& 
         {\makecell{Fishbone\\Diagram}} & {\makecell{Flow\\Chart}} & {\makecell{Foodchain}} & {\makecell{Gantt\\Chart}} & {\makecell{House\\Design\\Diagram}} & {\makecell{Kanban\\Board}} & {\makecell{Knowledge\\Graph}} & {\makecell{Matrix\\Diagram}} & \makecell{Mockups} & \makecell{Organizational\\Chart}\\
         \midrule
         BLIP-2~\cite{li2023blip2} &Flan-T5-XL &37.8 & 36.2 & 34.9 & 32.6 & 25.0 & \underline{48.8} & 39.5 & 30.0 & 43.8 & 39.1 \\
         InstructBLIP~\cite{dai2023instructblip} &Flan-T5-XL &37.8 & 38.3 & 41.9 & 34.9 & 15.9 & 36.6 & 31.6 & 30.0 & 46.9 & 41.3\\
         InstructBLIP Vicuna~\cite{dai2023instructblip} &Vicuna-7B & 26.7 & 40.4 & 27.9 & 25.6 & 18.2 & 26.8 & 52.6 & 22.5 & 46.9 & 34.8 \\
         LLaVA~\cite{liu2023visual_llava} &LLaMA-7B & 15.6 & 40.4 & 41.9 & \underline{39.5} & 29.6 & 22.0 & 55.3 & 15.0 & 53.1 & 30.4\\
         MiniGPT-4~\cite{zhu2023minigpt4} &Vicuna-7B & 28.9 & 36.2 & 27.9 & 32.6 & 25.0 & 36.6 & 36.8 & 20.0 & 50.0 & 41.3 \\
         VPGTrans~\cite{2023vpgtrans} &LLaMA-7B & 26.7 & 34.0 & 20.9 & 27.9 & 15.9 & 46.3 & 55.3 & 17.5 & 37.5 & 37.0 \\
         MultiModal-GPT~\cite{gong2023multimodalgpt} &Vicuna-7B &26.7 & 46.8 & 30.2 & 37.2 & 25.0 & 36.6 & 55.3 & 17.5 & 37.5 & 39.1\\
         Otter~\cite{li2023otter} &LLaMA-7B & 28.9 & 44.7 & 27.9 & 32.6 & 20.5 & 31.7 & 55.3 & 15.0 & 37.5 & 37.0\\
         OpenFlamingo~\cite{openflamingo} &LLaMA-7B & 26.7 & 46.8 & 30.2 & 37.2 & 25.0 & 36.6 & 55.3 & 17.5 & 37.5 & 39.1 \\
         LLaMA-Adapter V2~\cite{gao2023llamaadapterv2} &LLaMA-7B &24.4 & 38.3 & 23.3 & \underline{39.5} & 13.6 & 43.9 & 44.7 & 20.0 & 50.0 & 34.8 \\
         GVT~\cite{wang2023gvt} &Vicuna-7B & 24.4 & 38.3 & 23.3 & 32.6 & 22.7 & 29.3 & 44.7 & 17.5 & 43.8 & 34.8\\
         mPLUG-Owl~\cite{ye2023mplugowl} &LLaMA-7B &20.0 & 46.8 & 30.2 & 34.9 & 22.7 & 36.6 & 47.4 & 17.5 & 40.6 & 39.1\\
         Qwen-VL~\cite{bai2023qwen} & Qwen-7B & 37.8 & 44.7 & 39.5 & 34.9 & 34.1 & 41.5 & 44.7 & 30.0 & 56.3 & 47.8 \\
         Qwen-VL-Chat~\cite{bai2023qwen} & Qwen-7B &33.3 & 44.7 & 41.9 & 34.9 & 36.4 & 29.3 & 50.0 & 27.5 & 56.3 & 52.2\\
         LLaVA-1.5~\cite{liu2023llava1.5} &Vicuna-7B  & 37.8 & 51.1 & 34.9 & 32.6 & 20.5 & 31.7 & 63.2 & 22.5 & 62.5 & 39.1\\
         IDEFICS-9B-Instruct~\cite{laurencon2023obelics} & LLaMA-7B &24.4 & 44.7 & 30.2 & 32.6 & 15.9 & 36.6 & 50.0 & 22.5 & 43.8 & 34.8\\
         InternLM-Xcomposer-VL~\cite{zhang2023internlm} & InternLM-7B & 37.8 & 53.2 & 58.1 & 27.9 & 38.6 & 24.4 & 42.1 & 37.5 & 43.8 & 37.0\\
         VideoChat~\cite{li2023videochat} &Vicuna-7B & 24.4 & 31.9 & 32.6 & 23.3 & 20.5 & 31.7 & 39.5 & 17.5 & 37.5 & 43.5 \\
         Video-ChatGPT~\cite{maaz2023videochatgpt} &LLaMA-7B & 31.1 & 36.2 & 27.9 & 32.6 & 20.5 & 26.8 & 36.8 & 15.0 & 53.1 & 32.6\\
         Valley~\cite{luo2023valley} &LLaMA-13B & 22.2 & 38.3 & 23.3 & 34.9 & 20.5 & 36.6 & 44.7 & 17.5 & 53.1 & 39.1\\
         Emu~\cite{sun2023emu} & LLaMA-13B & 31.1 & 38.3& 37.2& \underline{39.5}& 25.0& 26.8& 57.9& 12.5& 56.3& 37.0\\
         NExt-GPT~\cite{wu2023nextgpt} & Vicuna-7B & 37.8& 29.8& 30.2& 32.6& 25.0& 29.3& 29.0& 20.0& 53.1& 19.6\\
         SEED-LLaMA~\cite{ge2023making} & LLaMA2-Chat-13B & 28.9& 42.6& 39.5& 34.9& 15.9& 31.7& 52.6& 15.0& 59.4& 37.0\\
         CogVLM~\cite{wang2023cogvlm} & Vicuna-7B & 28.9 & 44.7 & 34.9 & \underline{39.5} & 9.1 & 34.2 & 52.6 & 25.0 & 56.3 & 28.3 \\
         InternLM-Xcomposer-VL2~\cite{dong2024internlm2} & InternLM2-7B & \underline{42.2} & 53.2 & 65.1 & \bf{44.2} & 40.9 & 31.7 & 60.5 & 37.5 & 59.4 & 34.8\\
         InternLM-Xcomposer-VL2-4bit~\cite{dong2024internlm2} & InternLM2-7B & 24.4 & 51.1 & 46.5 & 34.9 & 34.1 & 36.6 & 50.0 & 35.0 & 40.6 & 21.7\\
         LLaVA-Next~\cite{liu2024llava_next} & Vicuna-7B & 28.9 & 46.8 & 34.9 & 34.9 & 31.8 & 36.6 & 47.4 & 30.0 & 46.9 & 39.1\\
         Yi-VL~\cite{yi_vl_6b} & Yi-6B & 28.9 & 36.2 & 37.2 & 25.6 & 27.3 & 12.2 & 47.4 & \underline{40.0} & 43.8 & 41.3 \\
         SPHINX-v2-1k~\cite{gao2024sphinx} &LLaMA2-7B &\underline{42.2} & 53.2 & 34.9 & 20.9 & 43.2 & 29.3 & 50.0 & \underline{40.0} & \bf{65.6} & \bf{60.9}\\
         mPLUG-Owl2~\cite{ye2023mplug2} &LLaMA2-7B & 31.1 & 44.7 & 30.2 & 34.9 & 25.0 & 41.5 & 52.6 & 20.0 & 43.8 & 41.3\\
         SEED-X~\cite{ge2024seedx} & LLaMA2-13B & 37.8 & 51.1 & 53.5 & 30.2 & 34.1 & 36.6 & 55.3 & \underline{40.0} & \bf{65.6} & 47.8 \\
         GPT-4V~\cite{2023GPT4VisionSC} & - & \bf{51.1} & \underline{66.0} & \underline{67.4} & 30.2 & \underline{45.5} & 40.0 & \bf{76.3} & \bf{45.0} & {61.3} & 42.2 \\
         Gemini-Pro-Vision~\cite{team2023gemini}  & - & 34.2 & \bf{66.7} & \bf{70.0} & 27.5 & \bf{46.5} & \bf{57.5} & \underline{65.0} & 27.6 & 51.6 & \underline{52.3} \\
         Claude-3-Opus~\cite{claude3family}  & - & 35.6 & 59.6 & \underline{67.4} & 32.6 & 38.6 & 22.0 & 57.9 & 37.5 & 48.4 & 19.6\\
         \bottomrule
    \end{tabular}
    }
   }
\end{table*}

\vspace{20pt}
\begin{table*}[]
    \vspace{5pt}
    \centering
    \caption{Evaluation results of various MLLMs in different types of ``Charts'' (Part 2) in SEED-Bench-2-plus. The best (second best) is in bold (underline).} \label{tab:chart_performance_part2}
    \vspace{3pt}
    {
    \resizebox{\textwidth}{!}{
    \begin{tabular}{cccccccccccc}
         \toprule
         \multirow{1}{*}{Model} & \multirow{1}{*}{Language Model}& \makecell{Pert\\Chart}& \makecell{Sankey\\Diagram}& \makecell{Spider\\Diagram}& \makecell{Storyboard} & \makecell{Swot\\Analysis}& \makecell{Timeline} & \makecell{Tree\\Diagram}& \makecell{Venn\\Diagramm}& \makecell{Wireframes} & \makecell{Word \\ Cloud}\\
         \midrule
         BLIP-2~\cite{li2023blip2} &Flan-T5-XL & 23.3 & \underline{40.0} & 22.2 & 6.7 & 26.3 & 50.0 & 32.4 & 37.0 & 33.3 & 43.2\\
         InstructBLIP~\cite{dai2023instructblip} &Flan-T5-XL & 20.9 & 36.7 & 22.2 & 17.8 & 23.7 & 36.7 & 29.4 & 30.4 & 25.0 & 40.9\\
         InstructBLIP Vicuna~\cite{dai2023instructblip} &Vicuna-7B & 23.3 & 20.0 & 31.1 & 13.3 & 31.6 & 46.7 & 17.7 & 23.9 & 41.7 & 36.4\\
         LLaVA~\cite{liu2023visual_llava} &LLaMA-7B & 27.9 & 20.0 & 24.4 & 11.1 & 31.6 & 43.3 & 29.4 & 17.4 & 36.1 & 25.0\\
         MiniGPT-4~\cite{zhu2023minigpt4} &Vicuna-7B & 34.9 & 26.7 & 20.0 & 11.1 & 29.0 & 36.7 & 29.4 & 21.7 & 38.9 & 34.1\\
         VPGTrans~\cite{2023vpgtrans} &LLaMA-7B & 34.9 & 13.3 & 31.1 & 17.8 & 26.3 & 46.7 & 14.7 & 26.1 & 41.7 & 31.8 \\
         MultiModal-GPT~\cite{gong2023multimodalgpt} &Vicuna-7B & 27.9 & 16.7 & 20.0 & 11.1 & 23.7 & 40.0 & 23.5 & 30.4 & 36.1 & 29.6\\
         Otter~\cite{li2023otter} &LLaMA-7B & 32.6 & 20.0 & 22.2 & 6.7 & 23.7 & 46.7 & 26.5 & 26.1 & 33.3 & 27.3\\
         OpenFlamingo~\cite{openflamingo} &LLaMA-7B  & 27.9 & 16.7 & 20.0 & 11.1 & 23.7 & 40.0 & 23.5 & 30.4 & 36.1 & 29.6\\
         LLaMA-Adapter V2~\cite{gao2023llamaadapterv2} &LLaMA-7B & 25.6 & 16.7 & 24.4 & 8.9 & 18.4 & 43.3 & 32.4 & 15.2 & 52.8 & 38.6 \\
         GVT~\cite{wang2023gvt} &Vicuna-7B & 25.6 & 20.0 & 17.8 & 22.2 & 36.8 & 40.0 & 41.2 & 17.4 & 36.1 & 27.3\\
         mPLUG-Owl~\cite{ye2023mplugowl} &LLaMA-7B & 34.9 & 6.7 & 26.7 & 15.6 & 29.0 & 43.3 & 26.5 & 26.1 & 36.1 & 31.8\\
         Qwen-VL~\cite{bai2023qwen} & Qwen-7B & 27.9 & 26.7 & 20.0 & 42.2 & 42.1 & 50.0 & 44.1 & 23.9 & 52.8 & 31.8\\
         Qwen-VL-Chat~\cite{bai2023qwen} & Qwen-7B & 37.2 & 20.0 & 24.4 & 55.6 & 34.2 & 46.7 & 41.2 & 15.2 & 41.7 & 27.3\\
         LLaVA-1.5~\cite{liu2023llava1.5} &Vicuna-7B & 27.9 & 20.0 & 24.4 & 40.0 & 39.5 & 53.3 & 29.4 & 26.1 & 44.4 & 38.6\\
         IDEFICS-9B-Instruct~\cite{laurencon2023obelics} & LLaMA-7B & 27.9 & 23.3 & 22.2 & 22.2 & 26.3 & 50.0 & 26.5 & 26.1 & 36.1 & 31.8\\
         InternLM-Xcomposer-VL~\cite{zhang2023internlm} & InternLM-7B & 37.2 & 30.0 & 37.8 & 31.1 & 36.8 & 50.0 & 32.4 & 39.1 & 52.8 & 50.0 \\
         VideoChat~\cite{li2023videochat} &Vicuna-7B & 25.6 & 23.3 & 26.7 & 8.9 & 29.0 & 33.3 & 26.5 & 15.2 & 36.1 & 34.1\\
         Video-ChatGPT~\cite{maaz2023videochatgpt} &LLaMA-7B & 23.3 & 20.0 & 26.7 & 28.9 & 23.7 & 40.0 & 38.2 & 19.6 & 41.7 & 31.8\\
         Valley~\cite{luo2023valley} &LLaMA-13B & 23.3 & 23.3 & 22.2 & 20.0 & 26.3 & 26.7 & 26.5 & 23.9 & 36.1 & 29.6\\
         Emu~\cite{sun2023emu} & LLaMA-13B & 27.9& 16.7& 22.2& 8.9& 36.8& 36.7& 35.3& 26.1& 44.4 & 38.6\\
         NExt-GPT~\cite{wu2023nextgpt} & Vicuna-7B & 20.9& 23.3& 20.0& 13.3& 34.2& 13.3& 26.5& 21.7& 25.0& 25.0\\
         SEED-LLaMA~\cite{ge2023making} & LLaMA2-Chat-13B & 34.9& 13.3& 33.3& 17.8& 36.8& 40.0& 23.5& 32.6& 33.3& 29.6\\
         CogVLM~\cite{wang2023cogvlm} & Vicuna-7B & 25.6 & 30.0 & 22.2 & 24.4 & 39.5 & 46.7 & 20.6 & 23.9 & 38.9 & 38.6\\
         InternLM-Xcomposer-VL2~\cite{dong2024internlm2} & InternLM2-7B & 32.6 & 33.3 & 51.1 & \bf{68.9} & 52.6 & 53.3 & \underline{52.9} & 56.5 & \underline{66.7} & \underline{52.3}\\
         InternLM-Xcomposer-VL2-4bit~\cite{dong2024internlm2} & InternLM2-7B & 27.9 & 30.0 & 51.1 & 40.0 & 44.7 & 36.7 & 32.4 & 23.9 & 38.9 & 47.7\\
         LLaVA-Next~\cite{liu2024llava_next} & Vicuna-7B & 25.6 & 30.0 & 26.7 & 42.2 & 34.2 & 53.3 & 44.1 & 21.7 & 38.9 & 43.2\\
         Yi-VL~\cite{yi_vl_6b} & Yi-6B & 23.3 & 16.7 & 26.7 & 33.3 & 42.1 & 26.7 & 29.4 & 23.9 & 47.2 & 38.6\\
         SPHINX-v2-1k~\cite{gao2024sphinx} &LLaMA2-7B & 39.5 & 16.7 & 35.6 & 53.3 & 42.1 & 56.7 & 32.4 & 32.6 & 50.0 & 36.4\\
         mPLUG-Owl2~\cite{ye2023mplug2} &LLaMA2-7B & 20.9 & 30.0 & 31.1 & 26.7 & 36.8 & 36.7 & 26.5 & 23.9 & 41.7 & 34.1\\
         SEED-X~\cite{ge2024seedx} & LLaMA2-13B & 37.2 & \bf{50.0} & 46.7 & 53.3 & \underline{65.8} & 53.3 & 38.2 & 34.8 & \bf{69.4} & 50.0 \\
         GPT-4V~\cite{2023GPT4VisionSC} & - & \underline{39.5} & {40.0} & \bf{62.2} & \underline{65.9} & \underline{65.8} & \bf{66.7} & \bf{61.8} & \bf{60.9} & 55.6 & \bf{55.8}\\
         Gemini-Pro-Vision~\cite{team2023gemini}  & -  & \bf{41.5} & \underline{48.2} & \underline{57.5} & 56.1 & \bf{67.7} & \underline{62.1} & 46.4 &\underline{58.5} & {58.6} & 50.0\\
         Claude-3-Opus~\cite{claude3family} & - & 37.2 & 33.3 & 55.6 & 53.5 & 60.5 & 43.3 & 50.0 & 41.3 & 36.1 & 45.5 \\
         \bottomrule
    \vspace{4pt}
    \end{tabular}
    }
   }
\end{table*}

Our SEED-Bench-2-Plus benchmark incorporates 2K multiple-choice questions, all of which are accompanied by accurate human annotations and span three broad categories including Chats, Maps and Webs. In this section, we first the broad categories of SEED-Bench-2-Plus in Sec.~\ref{sec:evaluate_dimensions}. We then introduce the data source in Sec.~\ref{sec:data_source}, and finally, we describe the evaluation strategy for MLLMs to answer multiple-choice questions in Sec.~\ref{sec:strategy}.

\subsection{Broad Categories}\label{sec:evaluate_dimensions}
To thoroughly evaluate the capabilities of MLLMs in comprehending text-rich data, SEED-Bench-2-Plus encompasses 3 broad categories (see Figure.~\ref{fig:text_rich_example}), which can be further divided into 63 data types (see Figure.~\ref{fig:text_rich_overview}).

\noindent\textbf{Charts.} This category pertains to the information contained within the chart image. In this task, the model is required to understand the specific semantics of each chart type, extract relevant information, and answer questions based on the context and spatial relationships. 

\noindent\textbf{Maps.} This category corresponds to the information present in the map image. The model is expected to identify symbols, text, and spatial relationships, and use this information to answer questions that often require geographical or domain-specific knowledge.

\noindent\textbf{Webs.} In this category, the model needs to understand the layout and design of different websites, extract relevant information from various elements, and answer questions that may relate to the website's content, functionality, or design based on the given website screenshot.

\subsection{Data Source}\label{sec:data_source}
To construct a comprehensive benchmark that encapsulates a variety of evaluation scenarios, it is imperative to amass an extensive collection of data. This data should predominantly include images that are rich in textual information, thereby creating a text-rich dataset.

\noindent\textbf{Charts.} To obtain chart data with rich text information, we manually gather 20 types of charts from the Internet. These types encompass Fishbone Diagram, Flow Chart, Foodchain, Gantt Chart, House Design Diagram, Kanban Board, Knowledge Graph, Matrix Diagram, Mockups, Organizational Chart, Pert Chart, Sankey Diagram, Spider Diagram, Storyboard, Swot Analysis, Timeline, Tree Diagram, Venn Diagram, Wireframes, and Word Cloud. Specifically, we employ GPT-4V~\cite{2023GPT4VisionSC} to generate corresponding questions and use human annotators to enhance the quality of the questions and corresponding options.

\noindent\textbf{Maps.} As for map data rich in text information, we manually collect 25 types of maps from the Internet. These types include Astronomical Map, Concept Map, Crime Map, Demographic Map, Environmental Map, Flow Map, Geologic Map, Heat Map, Historical Map, Infographics, Land Use Map, Migration Map, Mindmap, Nautical Map, Political Map, Resource Map, Road Map, Route Map, Site Map, Thematic Maps, Topographic Map, Transit Map, User Journey Map, Water Cycle, and Weather Map. Specifically, we utilize GPT-4V~\cite{2023GPT4VisionSC} to generate corresponding questions and employ human annotators to enhance the quality of the questions and corresponding options.

\noindent\textbf{Webs.} For web data rich in text information, we manually gather 18 types of screenshots from various websites on the Internet. These types include Airbnb, Amazon, Ebay, Github, Google Scholar Page, Homepage, Indeed, Linkedin, Papers, Poster, Project Page, Stackoverflow, Steam, Steam Game Setting, Tripadvisor, Twitter, Wikipedia, and Yelp. Specifically, we utilize GPT-4V~\cite{2023GPT4VisionSC} to generate corresponding questions and employ human annotators to enhance the quality of the questions and corresponding options.

\subsection{Evaluation Strategy}\label{sec:strategy}
\label{sec:strategy}
Different from MMBench~\cite{liu2023mmbench}, which employs ChatGPT to match a model's prediction to one of the choices in a multiple-choice question (achieving only an 87.0\% alignment rate), we adopt the answer ranking strategy~\cite{dai2023instructblip, brown2020gpt3, lin2021truthfulqa} to evaluate existing MLLMs with multiple-choice questions. Specifically, for each choice of a question, we compute the likelihood that an MLLM generates the content of this choice given the question. We select the choice with the highest likelihood as the model's prediction. Our evaluation strategy does not depend on the instruction-following capabilities of models to output ``A'', ``B'', ``C'' or ``D''. Also, this evaluation strategy eliminates the impact of the order of multiple-choice options on the model's performance.

\begin{table*}[]
    \centering
    \caption{Evaluation results of various MLLMs in different types of ``Maps'' (Part 1) in SEED-Bench-2-plus. The best (second best) is in bold (underline).} \label{tab:maps_performance_part1}
    \vspace{3pt}
    {
    \resizebox{\textwidth}{!}{
    \begin{tabular}{ccccccccccccccc}
         \toprule
         \multirow{1}{*}{Model} & \multirow{1}{*}{Language Model}& 
         {\makecell{Astronomical\\Maps}} & {\makecell{Concept\\Maps}} & {\makecell{Crime\\Maps}} & {\makecell{Demographic\\Maps}} & {\makecell{Environmental\\Maps}} & {\makecell{Flow\\Maps}} & {\makecell{Geologic\\Maps}} & {\makecell{Heat\\Maps}} & \makecell{Historical\\Maps} & \makecell{Infographic}& \makecell{Land\\Use\\Maps}& \makecell{Migration\\Maps}& \makecell{Mindmap}\\
         \midrule
         BLIP-2~\cite{li2023blip2} &Flan-T5-XL &20.0&39.6&34.6&31.6&14.7&23.3&18.5&26.1&26.7&43.1&27.8&23.7&22.0\\
         InstructBLIP~\cite{dai2023instructblip} &Flan-T5-XL &16.7&31.3&34.6&21.1&14.7&20.0&18.5&30.4&26.7&40.0&27.8&31.6&22.0\\
         InstructBLIP Vicuna~\cite{dai2023instructblip} &Vicuna-7B &20.0&37.5&38.5&42.1&29.4&33.3&14.8&34.8&40.0&32.3&11.1&44.7&29.3\\
         LLaVA~\cite{liu2023visual_llava} &LLaMA-7B &10.0&29.2&15.4&42.1&23.5&30.0&18.5&30.4&40.0&35.4&16.7&31.6&26.8\\
         MiniGPT-4~\cite{zhu2023minigpt4} &Vicuna-7B &13.3&22.9&26.9&36.8&26.5&16.7&18.5&34.8&53.3&33.9&19.4&31.6&29.3\\
         VPGTrans~\cite{2023vpgtrans} &LLaMA-7B &30.0&31.3&30.8&31.6&17.7&26.7&11.1&34.8&40.0&36.9&27.8&34.2&26.8\\
         MultiModal-GPT~\cite{gong2023multimodalgpt} &Vicuna-7B &26.7&39.6&34.6&42.1&26.5&30.0&14.8&34.8&53.3&38.5&16.7&44.7&24.4\\
         Otter~\cite{li2023otter} &LLaMA-7B &26.7&37.5&34.6&52.6&29.4&30.0&11.1&30.4&53.3&38.5&16.7&42.1&24.4\\
         OpenFlamingo~\cite{openflamingo} &LLaMA-7B &26.7&39.6&34.6&42.1&26.5&30.0&14.8&34.8&53.3&38.5&16.7&44.7&24.4\\
         LLaMA-Adapter V2~\cite{gao2023llamaadapterv2} &LLaMA-7B &16.7&33.3&42.3&47.4&29.4&23.3&14.8&17.4&40.0&33.9&22.2&\underline{47.4}&26.8\\
         GVT~\cite{wang2023gvt} &Vicuna-7B &16.7&31.3&30.8&36.8&29.4&33.3&29.6&34.8&33.3&29.2&16.7&42.1&34.2\\
         mPLUG-Owl~\cite{ye2023mplugowl} &LLaMA-7B &23.3&41.7&46.2&36.8&23.5&26.7&18.5&39.1&\underline{73.3}&36.9&19.4&42.1&29.3\\
         Qwen-VL~\cite{bai2023qwen} & Qwen-7B &13.3&31.3&34.6&31.6&29.4&40.0&14.8&34.8&40.0&53.9&33.3&39.5&26.8\\
         Qwen-VL-Chat~\cite{bai2023qwen} & Qwen-7B &20.0&43.8&30.8&26.3&26.5&36.7&22.2&17.4&40.0&55.4&36.1&29.0&26.8\\
         LLaVA-1.5~\cite{liu2023llava1.5} &Vicuna-7B  &20.0&35.4&46.2&36.8&26.5&26.7&25.9&26.1&46.7&38.5&16.7&39.5&39.0\\
         IDEFICS-9B-Instruct~\cite{laurencon2023obelics} & LLaMA-7B &16.7&29.2&38.5&31.6&23.5&40.0&22.2&26.1&46.7&41.5&22.2&\bf{50.0}&22.0\\
         InternLM-Xcomposer-VL~\cite{zhang2023internlm} & InternLM-7B &26.7&33.3&42.3&47.4&29.4&33.3&22.2&21.7&40.0&55.4&\underline{47.2}&42.1&{46.3}\\
         VideoChat~\cite{li2023videochat} &Vicuna-7B &16.7&33.3&19.2&36.8&23.5&23.3&22.2&30.4&33.3&30.8&22.2&36.8&24.4\\
         Video-ChatGPT~\cite{maaz2023videochatgpt} &LLaMA-7B &20.0&27.1&30.8&47.4&23.5&33.3&25.9&30.4&26.7&27.7&11.1&29.0&31.7\\
         Valley~\cite{luo2023valley} &LLaMA-13B &23.3&25.0&34.6&26.3&23.5&36.7&14.8&30.4&46.7&32.3&19.4&42.1&17.1\\
         Emu~\cite{sun2023emu} & LLaMA-13B & 20.0& 37.5& 46.2& 47.4& 26.5& 23.3& 22.2& 39.1& 46.7& 38.5& 22.2& 36.8& 26.8\\
         NExt-GPT~\cite{wu2023nextgpt} & Vicuna-7B & 30.0& 25.0& 15.4& 42.1& 23.5& 23.3& 33.3& 30.4& 13.3& 16.9& 25.0& 39.5& 24.4\\
         SEED-LLaMA~\cite{ge2023making} & LLaMA2-Chat-13B & 36.7& 37.5& 23.1& 42.1& 35.3& 20.0& 22.2& \underline{43.5}& 53.3& 36.9& 22.2& 44.7& 31.7\\
         CogVLM~\cite{wang2023cogvlm} & Vicuna-7B & 13.3&35.4&42.3&31.6&32.4&40.0&18.5&26.1&33.3&43.1&22.2&36.8&36.6\\
         InternLM-Xcomposer-VL2~\cite{dong2024internlm2} & InternLM2-7B & \bf{60.0} &56.3&\bf{65.4}&\bf{57.9}&35.3&\underline{50.0}&25.9&\bf{52.2}&46.7&64.6&30.6&34.2&{46.3}\\
         InternLM-Xcomposer-VL2-4bit~\cite{dong2024internlm2} & InternLM2-7B & 40.0&50.0&23.1&36.8&32.4&40.0&25.9&\underline{43.5}&60.0&55.4&38.9&36.8&43.9\\
         LLaVA-Next~\cite{liu2024llava_next} & Vicuna-7B & 23.3&35.4&34.6&31.6&29.4&33.3&22.2&34.8&33.3&40.0&5.6&29.0&26.8\\
         Yi-VL~\cite{yi_vl_6b} & Yi-6B & 20.0&54.2&23.1&26.3&17.7&26.7&14.8&30.4&40.0&44.6&22.2&31.6&41.5\\
         SPHINX-v2-1k~\cite{gao2024sphinx} & LLaMA2-7B &20.0&43.8&\underline{53.9}&42.1&\underline{38.2}&43.3&37.0&39.1&33.3&52.3&16.7&\underline{47.4}&39.0\\
         mPLUG-Owl2~\cite{ye2023mplug2} & LLaMA2-7B & 13.3&31.3&30.8&36.8&32.4&33.3&40.7&30.4&33.3&32.3&16.7&42.1&24.4\\
         SEED-X~\cite{ge2024seedx} & LLaMA2-13B & 43.3 & 47.9 & 30.8 & 36.8 & 32.4 & 43.3 & 29.6 & 34.8 & 60.0 & 58.5 & 38.9 & 39.5 & \bf{48.8} \\
         GPT-4V~\cite{2023GPT4VisionSC} & - & 55.2 &\bf{62.5} & 30.4 & \underline{52.9} & \bf{41.9} & 48.3 & \underline{42.3} & 28.6 & 66.7 & 67.2 & 32.3 & 46.0 & 41.5\\
         Gemini-Pro-Vision~\cite{team2023gemini}  & - & 46.2 & \underline{59.0} & 36.4 & 43.8 & 33.3 & \bf{53.6} & \bf{50.0} & 42.9 & \bf{76.9} & \bf{75.0} & \bf{52.2} & 42.4 & \underline{48.6}\\
         Claude-3-Opus~\cite{claude3family} & - & \underline{56.7} & 37.5 & 34.6 & 31.6 & 32.4 & 43.3 & 33.3 & 39.1 & 66.7 & \underline{67.7} & 44.4 & 36.8 & 43.9\\
         \bottomrule
    \end{tabular}
    }
   }
\end{table*}

\begin{table*}[]
    \centering
    \caption{Evaluation results of various MLLMs in different types of ``Maps'' (Part 2) in SEED-Bench-2-plus. The best (second best) is in bold (underline).} \label{tab:maps_performance_part2}
    \vspace{3pt}
    {
    \resizebox{\textwidth}{!}{
    \begin{tabular}{cccccccccccccc}
         \toprule
         \multirow{1}{*}{Model} & \multirow{1}{*}{Language Model} & \makecell{Nautical\\Maps} & \makecell{Political\\Maps}& \makecell{Resource\\Maps} & \makecell{Road\\Maps}& \makecell{Route\\Maps}& \makecell{Site\\Maps} & \makecell{Thematic \\ Maps} & \makecell{Topographic \\ Maps}  & \makecell{Transit \\ Maps}  & \makecell{User \\ Journey \\ Maps}  & \makecell{Water \\ Cycle} & \makecell{Weather \\ Maps}\\
         \midrule
         BLIP-2~\cite{li2023blip2} &Flan-T5-XL &16.7&50.0&39.1&36.4&37.5&38.2&12.5&44.0&34.5&23.7&39.0&22.6\\
         InstructBLIP~\cite{dai2023instructblip} &Flan-T5-XL &25.0&38.5&21.7&43.2&27.5&38.2&12.5&48.0&34.5&26.3&39.0&19.4\\
         InstructBLIP Vicuna~\cite{dai2023instructblip} &Vicuna-7B &16.7&53.9&13.0&\bf{59.1}&27.5&32.4&15.6&52.0&37.9&29.0&36.6&25.8\\
         LLaVA~\cite{liu2023visual_llava} &LLaMA-7B &25.0&42.3&30.4&45.5&25.0&29.4&15.6&32.0&48.3&26.3&31.7&25.8\\
         MiniGPT-4~\cite{zhu2023minigpt4} &Vicuna-7B &25.0&46.2&26.1&47.7&32.5&29.4&28.1&44.0&41.4&26.3&26.8&22.6\\
         VPGTrans~\cite{2023vpgtrans} &LLaMA-7B &25.0&50.0&39.1&43.2&20.0&26.5&34.4&36.0&24.1&23.7&46.3&32.3\\
         MultiModal-GPT~\cite{gong2023multimodalgpt} &Vicuna-7B &25.0&50.0&21.7&47.7&25.0&29.4&18.8&52.0&27.6&31.6&43.9&19.4\\
         Otter~\cite{li2023otter} &LLaMA-7B &25.0&50.0&34.8&45.5&27.5&29.4&25.0&52.0&27.6&21.1&41.5&16.1\\
         OpenFlamingo~\cite{openflamingo} &LLaMA-7B &25.0&50.0&21.7&47.7&25.0&29.4&18.8&52.0&27.6&31.6&43.9&19.4\\
         LLaMA-Adapter V2~\cite{gao2023llamaadapterv2} &LLaMA-7B &33.3&46.2&21.7&40.9&25.0&23.5&21.9&52.0&31.0&31.6&31.7&19.4\\
         GVT~\cite{wang2023gvt} &Vicuna-7B &25.0&34.6&21.7&45.5&30.0&23.5&25.0&36.0&41.4&18.4&29.3&25.8\\
         mPLUG-Owl~\cite{ye2023mplugowl} &LLaMA-7B &25.0&46.2&30.4&43.2&30.0&29.4&18.8&48.0&34.5&31.6&41.5&12.9\\
         Qwen-VL~\cite{bai2023qwen} & Qwen-7B &41.7&38.5&26.1&36.4&37.5&44.1&34.4&44.0&27.6&21.1&53.7&{48.4}\\
         Qwen-VL-Chat~\cite{bai2023qwen} & Qwen-7B &\bf{58.3}&38.5&34.8&29.6&27.5&\bf{50.0}&28.1&52.0&24.1&23.7&48.8&25.8\\
         LLaVA-1.5~\cite{liu2023llava1.5} &Vicuna-7B  &33.3&53.9&17.4&45.5&30.0&29.4&21.9&\bf{64.0}&27.6&36.8&46.3&19.4\\
         IDEFICS-9B-Instruct~\cite{laurencon2023obelics} & LLaMA-7B &33.3&42.3&34.8&52.3&22.5&20.6&21.9&36.0&31.0&23.7&41.5&16.1\\
         InternLM-Xcomposer-VL~\cite{zhang2023internlm} & InternLM-7B &16.7&\underline{73.1}&34.8&43.2&17.5&38.2&34.4&44.0&27.6&{44.7}&56.1&29.0\\
         VideoChat~\cite{li2023videochat} &Vicuna-7B &25.0&50.0&26.1&45.5&27.5&29.4&12.5&32.0&48.3&34.2&31.7&16.1\\
         Video-ChatGPT~\cite{maaz2023videochatgpt} &LLaMA-7B &25.0&38.5&30.4&40.9&32.5&26.5&28.1&32.0&41.4&21.1&29.3&19.4\\
         Valley~\cite{luo2023valley} &LLaMA-13B &33.3&34.6&26.1&29.6&27.5&20.6&12.5&32.0&27.6&26.3&34.2&12.9\\
         Emu~\cite{sun2023emu} & LLaMA-13B & 25.0& 42.3& 21.7& 52.3& 30.0& 32.4& 26.5& 18.8& \bf{60.0}& 37.9& 34.2& 41.5\\
         NExt-GPT~\cite{wu2023nextgpt} & Vicuna-7B & 25.0& 30.8& 30.4& 20.5& 30.0& 26.5& 23.5& 21.9& 24.0& \bf{48.3}& 29.0& 34.2\\
         SEED-LLaMA~\cite{ge2023making} & LLaMA2-Chat-13B & 33.3& 57.7& 30.4& 43.2& 25.0& 44.1& \underline{41.2}& 18.8& {52.0}& 27.6& 39.5& {48.8}\\
         CogVLM~\cite{wang2023cogvlm} & Vicuna-7B &25.0&69.2&17.4&43.2&25.0&23.5&25.0&\underline{60.0}&20.7&34.2&43.9&19.4\\
         InternLM-Xcomposer-VL2~\cite{dong2024internlm2} & InternLM2-7B &25.0&65.4&{43.5}&47.7&40.0&32.4&{40.6}&52.0&48.3&34.2&68.3&38.7\\
         InternLM-Xcomposer-VL2-4bit~\cite{dong2024internlm2} & InternLM2-7B &41.7&53.9&30.4&36.4&32.5&32.4&15.6&32.0&34.5&39.5&51.2&25.8\\
         LLaVA-Next~\cite{liu2024llava_next} & Vicuna-7B &41.7&38.5&26.1&50.0&32.5&35.3&25.0&48.0&48.3&34.2&41.5&22.6\\
         Yi-VL~\cite{yi_vl_6b} & Yi-6B &50.0&50.0&39.1&43.2&22.5&47.1&25.0&52.0&24.1&31.6&34.2&41.9\\
         SPHINX-v2-1k~\cite{gao2024sphinx} & LLaMA2-7B &\bf{58.3}&65.4&17.4&43.2&\underline{42.5}&\bf{50.0}&18.8&56.0&37.9&23.7&68.3&25.8\\
         mPLUG-Owl2~\cite{ye2023mplug2} & LLaMA2-7B &33.3&46.2&21.7&43.2&35.0&26.5&25.0&44.0&34.5&29.0&46.3&25.8\\
         SEED-X~\cite{ge2024seedx} & LLaMA2-13B & 41.7 & 61.5 & 17.4 & 40.9 & 35.0 & 44.1 & \bf{50.0} & 31.3 & \bf{60.0} & \bf{48.3} & 34.2 & \underline{53.7} \\
         GPT-4V~\cite{2023GPT4VisionSC} & - & 54.6 & \bf{76.9} &\underline{45.5} & 45.5 & \bf{48.6} & 44.1 & 37.5 & 52.2 & 41.4 & 43.2 & \bf{73.7} & 45.2\\
         Gemini-Pro-Vision~\cite{team2023gemini}  & - & 57.1 & 47.6 & 38.1 & \underline{54.1} & 33.3 & 43.5 & 32.1 & 50.0 & 30.0 & {45.7} & \underline{72.5} & {44.4}\\
         Claude-3-Opus~\cite{claude3family} & - & 33.3 & \underline{73.1} & \bf{47.8} & 40.9 & 32.5 & 41.2 & 38.2 & 21.9 & {56.0} & 34.5 & 44.7 & \bf{68.3}\\
         \bottomrule
    \end{tabular}
    }
   }
\end{table*}

\begin{table*}[]
    \centering
    \caption{Evaluation results of various MLLMs in different types of ``Webs'' (Part 1) in SEED-Bench-2-plus. The best (second best) is in bold (underline).} \label{tab:webs_performance_part1}
    \vspace{3pt}
    {
    \resizebox{\textwidth}{!}{
    \begin{tabular}{ccccccccccc}
         \toprule
         \multirow{1}{*}{Model} & \multirow{1}{*}{Language Model} & 
         {\makecell{Airbnb}} & {\makecell{Amazon}} & {\makecell{Ebay}} & {\makecell{Github}} & {\makecell{Google\\Scholar\\Page}} & {\makecell{Homepage}} & {\makecell{Indeed}} & {\makecell{Linkedin}} & \makecell{Papers}\\
         \bottomrule
         BLIP-2~\cite{li2023blip2} &Flan-T5-XL & 14.3 & 20.0 & 18.2 & 20.0 & 13.9 & 40.0 & 34.2 & 23.8 & 16.7 \\
         InstructBLIP~\cite{dai2023instructblip} &Flan-T5-XL &16.3 & 20.0 & 25.0 & 30.0 & 16.7 & 35.0 & 34.2 & 28.6 & 19.4 \\
         InstructBLIP Vicuna~\cite{dai2023instructblip} &Vicuna-7B & 12.2 & 24.4 & 22.7 & 30.0 & 19.4 & 42.5 & 34.2 & 23.8 & 36.1 \\
         LLaVA~\cite{liu2023visual_llava} &LLaMA-7B & 16.3 & 17.8 & 25.0 & 30.0 & 27.8 & 42.5 & 42.1 & 23.8 & 41.7\\
         MiniGPT-4~\cite{zhu2023minigpt4} &Vicuna-7B &16.3 & 22.2 & 25.0 & 30.0 & 27.8 & 47.5 & 31.6 & 16.7 & 33.3 \\
         VPGTrans~\cite{2023vpgtrans} &LLaMA-7B & 10.2 & 24.4 & 31.8 & 35.0 & 27.8 & 40.0 & 31.6 & 26.2 & 36.1\\
         MultiModal-GPT~\cite{gong2023multimodalgpt} &Vicuna-7B & 24.5 & 22.2 & 27.3 & 35.0 & 27.8 & 47.5 & 29.0 & 31.0 & 36.1\\
         Otter~\cite{li2023otter} &LLaMA-7B & 22.5 & 26.7 & 25.0 & 35.0 & 38.9 & 52.5 & 26.3 & 26.2 & 36.1 \\
         OpenFlamingo~\cite{openflamingo} &LLaMA-7B & 24.5 & 22.2 & 27.3 & 35.0 & 27.8 & 47.5 & 29.0 & 31.0 & 36.1\\
         LLaMA-Adapter V2~\cite{gao2023llamaadapterv2} &LLaMA-7B & 20.4 & 26.7 & 22.7 & 30.0 & 30.6 & 42.5 & 34.2 & 16.7 & 30.6\\
         GVT~\cite{wang2023gvt} &Vicuna-7B &18.4 & 20.0 & 20.5 & 30.0 & 33.3 & 42.5 & 39.5 & 21.4 & 38.9 \\
         mPLUG-Owl~\cite{ye2023mplugowl} &LLaMA-7B &16.3 & 24.4 & 25.0 & 35.0 & 30.6 & 45.0 & 31.6 & 26.2 & 30.6 \\
         Qwen-VL~\cite{bai2023qwen} & Qwen-7B & \underline{71.4} & \bf{51.1} & 63.6 & 60.0 & 47.2 & 47.5 & 65.8 & 52.4 & 41.7\\
         Qwen-VL-Chat~\cite{bai2023qwen} & Qwen-7B & \bf{77.6} & 46.7 & 63.6 & \bf{65.0} & 47.2 & 45.0 & \underline{68.4} & 57.1 & {47.2}\\
         LLaVA-1.5~\cite{liu2023llava1.5} &Vicuna-7B  & 26.5 & 28.9 & 43.2 & 45.0 & 25.0 & 37.5 & 42.1 & 42.9 & {47.2}\\
         IDEFICS-9B-Instruct~\cite{laurencon2023obelics} & LLaMA-7B &32.7 & 24.4 & 22.7 & 35.0 & 27.8 & 50.0 & 29.0 & 26.2 & 36.1 \\
         InternLM-Xcomposer-VL~\cite{zhang2023internlm} & InternLM-7B &34.7 & 42.2 & 47.7 & 60.0 & 25.0 & 52.5 & 57.9 & 42.9 & 44.4 \\
         VideoChat~\cite{li2023videochat} &Vicuna-7B & 10.2 & 22.2 & 25.0 & 35.0 & 22.2 & 45.0 & 31.6 & 14.3 & 33.3\\
         Video-ChatGPT~\cite{maaz2023videochatgpt} &LLaMA-7B &14.3 & 26.7 & 22.7 & 30.0 & 27.8 & 45.0 & 26.3 & 21.4 & 41.7 \\
         Valley~\cite{luo2023valley} &LLaMA-13B &28.6 & 20.0 & 22.7 & 25.0 & 33.3 & 42.5 & 26.3 & 26.2 & 41.7 \\
         Emu~\cite{sun2023emu} & LLaMA-13B & 35.5& 12.2& 24.4& 36.4& 35.0& 27.8& 47.5& 36.8& 33.3\\
         NExt-GPT~\cite{wu2023nextgpt} & Vicuna-7B & 16.1& 18.4& 26.7& 27.3& 30.0& 25.0& 25.0& 23.7& 16.7\\
         SEED-LLaMA~\cite{ge2023making} & LLaMA2-Chat-13B & 22.6& 22.5& 20.0& 34.1& 40.0& 22.2& 47.5& 34.2& 31.0\\
         CogVLM~\cite{wang2023cogvlm} & Vicuna-7B & 24.5 & 22.2 & 36.4 & 45.0 & 30.6 & 42.5 & 42.1 & 26.2 & {47.2}\\
         InternLM-Xcomposer-VL2~\cite{dong2024internlm2} & InternLM2-7B & 53.1 & \bf{51.1} & 50.0 & 65.0 & 38.9 & \bf{70.0} & 81.6 & 59.5 & {47.2}\\
         InternLM-Xcomposer-VL2-4bit~\cite{dong2024internlm2} & InternLM2-7B & 38.8 & 26.7 & 34.1 & 35.0 & 30.6 & 37.5 & 55.3 & 45.2 & 36.1 \\
         LLaVA-Next~\cite{liu2024llava_next} & Vicuna-7B & 28.6 & 26.7 & 40.9 & 55.0 & 22.2 & 45.0 & 52.6 & 40.5 & 44.4\\
         Yi-VL~\cite{yi_vl_6b} & Yi-6B & 40.8 & 28.9 & 27.3 & 45.0 & 30.6 & 35.0 & 42.1 & 38.1 & 41.7 \\
         SPHINX-v2-1k~\cite{gao2024sphinx} & LLaMA2-7B & 69.4 & 35.6 & \bf{65.9} & \bf{65.0} &{47.2} & \underline{65.0} & \bf{81.6} & 57.1 & 44.4 \\
         mPLUG-Owl2~\cite{ye2023mplug2} & LLaMA2-7B & 28.6 & 26.7 & 31.8 & 50.0 & 19.4 & 42.5 & 36.8 & 33.3 & 41.7 \\
         SEED-X~\cite{ge2024seedx} & LLaMA2-13B & 45.2 & 51.0 & 37.8 & 54.6 & 50.0 & 33.3 & 45.0 & \bf{73.7} & \underline{54.8}  \\
         GPT-4V~\cite{2023GPT4VisionSC} & - & 58.3 & 44.7 & 47.6 & 55.0 & \underline{51.4} & 43.6 & 66.7 & {62.5} & \bf{62.9}\\
         Gemini-Pro-Vision~\cite{team2023gemini} & - & 60.5 & 48.3 & \underline{55.3} & - & 40.0 & 47.4 & 60.0 & \underline{63.6} & 30.0 \\
         Claude-3-Opus~\cite{claude3family} & - & 23.3 & 34.7 & 37.8 & 20.5 & \bf{60.0} & 33.3 & 57.5 & 57.9 & 40.5\\
         \bottomrule
         \vspace{4pt}
    \end{tabular}
    }
   }
\end{table*}

\begin{table*}[]
    \centering
    \caption{Evaluation results of various MLLMs in different types of ``Webs'' (Part 2) in SEED-Bench-2-plus. The best (second best) is in bold (underline).} \label{tab:webs_performance_part2}
    \vspace{3pt}
    {
    \resizebox{\textwidth}{!}{
    \begin{tabular}{ccccccccccc}
         \toprule
         \multirow{1}{*}{Model} & \multirow{1}{*}{Language Model} & \makecell{Poster}& \makecell{Project\\Page}& \makecell{Stackoverflow}& \makecell{Steam}& \makecell{Steam\\Game\\Setting} & \makecell{Tripadvisior}& \makecell{Twitter} & \makecell{Wikipedia}& \makecell{Yelp} \\
         \bottomrule
         BLIP-2~\cite{li2023blip2} &Flan-T5-XL & 41.5 & 23.1 & 30.3 & 23.5 & 17.7 & 21.4 & 42.1 & 34.7 & 21.7\\
         InstructBLIP~\cite{dai2023instructblip} &Flan-T5-XL & 48.8 & 15.4 & 27.3 & 17.7 & 14.7 & 21.4 & 36.8 & 38.8 & 26.1\\
         InstructBLIP Vicuna~\cite{dai2023instructblip} &Vicuna-7B & 41.5 & 38.5 & 27.3 & 29.4 & 32.4 & 23.8 & 42.1 & 53.1 & 19.6\\
         LLaVA~\cite{liu2023visual_llava} &LLaMA-7B & 41.5 & 15.4 & 42.4 & 35.3 & 26.5 & 21.4 & 42.1 & 40.8 & 28.3\\
         MiniGPT-4~\cite{zhu2023minigpt4} &Vicuna-7B & 43.9 & 23.1 & 30.3 & 29.4 & 41.2 & 28.6 & 36.8 & 44.9 & 19.6\\
         VPGTrans~\cite{2023vpgtrans} &LLaMA-7B & 43.9 & 23.1 & 33.3 & 32.4 & 29.4 & 16.7 & 15.8 & 38.8 & 30.4\\
         MultiModal-GPT~\cite{gong2023multimodalgpt} &Vicuna-7B & 39.0 & 30.8 & 39.4 & 29.4 & 29.4 & 23.8 & 36.8 & 49.0 & 19.6\\
         Otter~\cite{li2023otter} &LLaMA-7B & 39.0 & 23.1 & 42.4 & 26.5 & 26.5 & 26.2 & 36.8 & 40.8 & 21.7\\
         OpenFlamingo~\cite{openflamingo} &LLaMA-7B & 39.0 & 30.8 & 39.4 & 29.4 & 29.4 & 23.8 & 36.8 & 49.0 & 19.6\\
         LLaMA-Adapter V2~\cite{gao2023llamaadapterv2} &LLaMA-7B & 46.3 & 30.8 & 42.4 & 38.2 & 32.4 & 16.7 & 47.4 & 36.7 & 30.4\\
         GVT~\cite{wang2023gvt} &Vicuna-7B & 31.7 & 30.8 & 30.3 & 35.3 & 29.4 & 28.6 & 36.8 & 40.8 & 17.4\\
         mPLUG-Owl~\cite{ye2023mplugowl} &LLaMA-7B & 41.5 & 30.8 & 36.4 & 29.4 & 35.3 & 23.8 & 36.8 & 44.9 & 28.3\\
         Qwen-VL~\cite{bai2023qwen} & Qwen-7B & 61.0 & {76.9} & 69.7 & 35.3 & \underline{64.7} & 59.5 & 73.7 & 40.8 & 54.4\\
         Qwen-VL-Chat~\cite{bai2023qwen} & Qwen-7B & 48.8 & 69.2 & \bf{72.7} & 38.2 & \bf{67.7} & \underline{64.3} & 73.7 & 44.9 & \underline{56.5}\\
         LLaVA-1.5~\cite{liu2023llava1.5} &Vicuna-7B & 58.5 & 53.9 & 51.5 & 35.3 & 58.8 & 31.0 & 52.6 & 42.9 & 21.7\\
         IDEFICS-9B-Instruct~\cite{laurencon2023obelics} & LLaMA-7B & 36.6 & 23.1 & 39.4 & 41.2 & 35.3 & 31.0 & 36.8 & 46.9 & 28.3\\
         InternLM-Xcomposer-VL~\cite{zhang2023internlm} & InternLM-7B  & 53.7 & 38.5 & 45.5 & 38.2 & 35.3 & 31.0 & 63.2 & 36.7 & 34.8\\
         VideoChat~\cite{li2023videochat} &Vicuna-7B & 36.6 & 30.8 & 33.3 & 29.4 & 35.3 & 31.0 & 52.6 & 40.8 & 10.9\\
         Video-ChatGPT~\cite{maaz2023videochatgpt} &LLaMA-7B & 41.5 & 38.5 & 36.4 & 29.4 & 32.4 & 31.0 & 42.1 & 38.8 & 21.7\\
         Valley~\cite{luo2023valley} &LLaMA-13B & 29.3 & 38.5 & 45.5 & 32.4 & 32.4 & 23.8 & 52.6 & 36.7 & 23.9\\
         Emu~\cite{sun2023emu} & LLaMA-13B & 38.9& 51.2& 38.5& 33.3& 24.2& 21.4& 42.1& {59.2}& 23.9\\
         NExt-GPT~\cite{wu2023nextgpt} & Vicuna-7B & 27.8& 31.7& 38.5& 36.4& 21.2& 23.8& 36.8& 26.5& 21.7\\
         SEED-LLaMA~\cite{ge2023making} & LLaMA2-Chat-13B & 38.9& 46.3& 15.4& 45.5& 30.3& 26.2& 31.6& 44.9& 26.1\\
         CogVLM~\cite{wang2023cogvlm} & Vicuna-7B & 46.3 & 30.8 & 33.3 & 29.4 & 50.0 & 23.8 & 47.4 & 32.7 & 26.1\\
         InternLM-Xcomposer-VL2~\cite{dong2024internlm2} & InternLM2-7B & 63.4 & \bf{84.6} & 63.6 & 50.0 & 41.2 & 57.1 & \bf{84.2} & 51.0 & 52.2\\
         InternLM-Xcomposer-VL2-4bit~\cite{dong2024internlm2} & InternLM2-7B & 43.9 & 38.5 & 27.3 & 50.0 & 38.2 & 23.8 & 68.4 & 32.7 & 21.7\\
         LLaVA-Next~\cite{liu2024llava_next} & Vicuna-7B & 51.2 & 61.5 & 48.5 & 33.3 & 55.9 & 33.3 & 63.2 & 42.9 & 28.3\\
         Yi-VL~\cite{yi_vl_6b} & Yi-6B & 51.2 & 30.8 & 42.4 & 35.3 & 35.3 & 38.1 & 63.2 & 36.7 & 26.1\\
         SPHINX-v2-1k~\cite{gao2024sphinx} & LLaMA2-7B & 58.5 & {76.9} & 57.6 & \underline{63.6} & \underline{64.7} & \bf{69.1} & 79.0 & {57.1} & \bf{58.7}\\
         mPLUG-Owl2~\cite{ye2023mplug2} & LLaMA2-7B & 41.5 & 15.4 & 33.3 & 50.0 & 38.2 & 14.3 & 57.9 & 42.9 & 23.9\\
         SEED-X~\cite{ge2024seedx} & LLaMA2-13B & 36.1 & 73.2 & 69.2 & 42.4 & 57.6 & 54.8 & 79.0 & 46.9 & \underline{56.5} \\
         GPT-4V~\cite{2023GPT4VisionSC} & - & \underline{68.3} & \underline{83.3} & 51.5 &\bf{70.6} & 42.4 & 40.0 & \underline{83.3} & \bf{65.3} & 47.6\\
         Gemini-Pro-Vision~\cite{team2023gemini} & - & \bf{82.4} & 75.0 & \underline{70.4} & 36.8 & 57.7 & 44.4 & 81.8 & 35.3 & 54.8\\
         Claude-3-Opus~\cite{claude3family} & - & 41.7 & 58.5 & 61.5 & 39.4 & 57.6 & 40.5 & 42.1 & \underline{63.3} & 39.1 \\
         \bottomrule
         \vspace{4pt}
    \end{tabular}
    }
   }
\end{table*}
\section{Evaluation Results}

\subsection{Models}
We evaluate a total of 31 open-source MLLMs including BLIP-2~\cite{li2023blip2}, InstructBLIP~\cite{dai2023instructblip}, InstructBLIP Vicuna~\cite{dai2023instructblip}, LLaVA~\cite{liu2023visual_llava}, MiniGPT-4~\cite{zhu2023minigpt4}, VPGTrans~\cite{2023vpgtrans},  MultiModal-GPT~\cite{gong2023multimodalgpt}, Otter~\cite{li2023otter}, OpenFlamingo~\cite{openflamingo}, LLaMA-Adapter V2~\cite{gao2023llamaadapterv2}, GVT~\cite{wang2023gvt}, mPLUG-Owl~\cite{ye2023mplugowl}, Qwen-VL~\cite{bai2023qwen}, Qwen-VL-Chat~\cite{bai2023qwen}, LLaVA1.5~\cite{liu2023llava1.5}, IDEFICS-9B-Instruct~\cite{laurencon2023obelics}, InternLM-Xcomposer-VL~\cite{zhang2023internlm}, VideoChat~\cite{li2023videochat}, Video-ChatGPT~\cite{maaz2023videochatgpt}, Valley~\cite{luo2023valley}, Emu~\cite{sun2023emu}, NExt-GPT~\cite{wu2023nextgpt}, SEED-LLaMA~\cite{ge2023making}, CogVLM~\cite{wang2023cogvlm}, InternLM-Xcomposer-VL-2~\cite{dong2024internlm2}, InternLM-Xcomposer-VL-2-4bit~\cite{dong2024internlm2}, LLaVA-Next~\cite{liu2024llava_next}, Yi-VL~\cite{yi_vl_6b}, SPHINX-v2-1k~\cite{gao2024sphinx}, mPLUG-Owl2~\cite{ye2023mplug2}, SEED-X~\cite{ge2024seedx} (We evaluate the general instruction-tuned model SEED-X-I), and 3 closed-source MLLMs including GPT-4V~\cite{2023GPT4VisionSC}, Gemini-Pro-Vision~\cite{team2023gemini}, and Claude-3-Opus~\cite{claude3family}, based on their official implementations. 
It is important to note that for Gemini-Pro-Vision, we only report task performance when the model responds to over half of the valid data in the task.

\subsection{Main Results} 
The evaluation results for various MLLMs across different categories of SEED-Bench-2-Plus are presented in Table~\ref{tab:performance}. The detailed results of 63 types are presented in Table~\ref{tab:chart_performance_part1}, Table~\ref{tab:chart_performance_part2}, Table~\ref{tab:maps_performance_part1}, Table~\ref{tab:maps_performance_part2}, Table~\ref{tab:webs_performance_part1}, and Table~\ref{tab:webs_performance_part2}, respectively. Notably, GPT-4V surpasses a significant number of MLLMs, demonstrating superior performance across most evaluation types.

To provide a more comprehensive overview of model capabilities across different categories, we have visualized the ranking of each model within each category in Figure~\ref{fig:rank}. Here, darker hues represent higher ranks. The top-performing MLLM, GPT-4V, exhibits competitive results across different categories.

\subsection{Observations}
Through a comprehensive and objective evaluation of various MLLMs across different text-rich scenarios in SEED-Bench-2-Plus, we have gleaned valuable insights that can guide future research efforts.

\noindent\textbf{Comprehending text-rich data proves to be more complex.} The majority of MLLMs achieve lower results on text-rich data, with the average accuracy rate being less than 40\%. Considering the potential of MLLMs as multimodal agents, particularly website agents, a crucial capability is to analyze a variety of websites and generate accurate responses. The unsatisfactory results indicate that significant advancements are required before MLLMs can be effectively utilized as multimodal agents. This underlines the complexity of comprehending text-rich data and highlights the need for further research and development in this area to enhance the proficiency of MLLMs in text-rich scenarios.

\noindent\textbf{Maps are more difficult for text-rich comprehension.} Maps are inherently complex and multidimensional, frequently featuring multiple overlapping layers of information. They not only display geographical details but also contain various symbols, colors, and texts to represent different types of information, such as topographical features, political boundaries, and points of interest. This complexity can pose challenges for models to accurately interpret and understand the full context of maps, compared with charts and webs.

\noindent\textbf{The performance of MLLMs varies significantly across different data types.} Different data types, such as knowledge graph and matrix diagram, which both fall under the category of charts, differ considerably in their complexity and structure. Some data types may require more sophisticated understanding and processing capabilities than others due to their inherent characteristics. As such, it is reasonable to observe a performance discrepancy of models across different data types. This finding underscores the need for MLLMs to be robust and adaptable, capable of handling a diverse range of data types in text-rich scenarios.

\section{Conclusion}
In this study, we present SEED-Bench-2-Plus, an extensive benchmark specifically designed for evaluating Multimodal Large Language Models (MLLMs) in text-rich scenarios. SEED-Bench-2-Plus comprises 2K multiple-choice questions, each annotated by humans, and span across 3 brod categories across 63 data types. We carry out a comprehensive evaluation of 31 noteworthy open-source MLLMs and 3 closed-source MLLMs, examining and comparing their performances to derive valuable insights that can guide future research. Serving as a valuable supplement to SEED-Bench-2~\cite{li2023seed2}, both the dataset and the evaluation code of SEED-Bench-2-Plus are made publicly available. Additionally, we consistently maintain a leaderboard to facilitate the advancements in the field of text-rich visual comprehension with MLLMs.

{
    \small
    \bibliographystyle{ieeenat_fullname}
    \bibliography{main}
}


\end{document}